\documentclass[letterpaper, 10pt, conference]{ieeeconf}      % Use this line for a4 paper

\IEEEoverridecommandlockouts                              % This command is only needed if 
\usepackage{graphics} % for pdf, bitmapped graphics files
\usepackage{epsfig} % for postscript graphics files
\usepackage{mathptmx} % assumes new font selection scheme installed
\usepackage{times} % assumes new font selection scheme installed
\usepackage{amsmath} % assumes amsmath package installed
\usepackage{amssymb}  % assumes amsmath package installed
\usepackage{graphicx}
\usepackage{booktabs}
\usepackage{cite}
\usepackage{hyperref}
\usepackage{multicol}
\usepackage{multirow}
\usepackage[dvipsnames]{xcolor}
\usepackage{colortbl}
\usepackage{diagbox}
\title{\LARGE \bf
CooPre: Cooperative Pretraining for V2X Cooperative Perception
}

\author{Seth Z. Zhao$^{1}$,  Hao Xiang$^{1}$, Chenfeng Xu$^{2}$,  Xin Xia$^{1}$,  Bolei Zhou$^{1}$, Jiaqi Ma$^{1 \dagger}$
\thanks{$^{\dagger}$ Corresponding Author. \tt \small{jiaqima@ucla.edu}. }
\thanks{$^{1}$ Seth Z. Zhao, Hao Xiang, Xin Xia, Bolei Zhou, and Jiaqi Ma are affiliated with the University of California, Los Angeles. \tt \small{sethzhao506@g.ucla.edu}.}
\thanks{$^{2}$ Chenfeng Xu is affiliated with the University of California, Berkeley.}}

\begin{document}
\maketitle
\thispagestyle{empty}
\pagestyle{empty}

%%%%%%%%%%%%%%%%%%%%%%%%%%%%%%%%%%%%%%%%%%%%%%%%%%%%%%%%%%%%%%%%%%%%%%%%%%%%%%%%
\begin{abstract}
Existing Vehicle-to-Everything (V2X) cooperative perception methods rely on accurate multi-agent 3D annotations. Nevertheless, it is time-consuming and expensive to collect and annotate real-world data, especially for V2X systems. In this paper, we present a self-supervised learning framwork for V2X cooperative perception, which utilizes the vast amount of unlabeled 3D V2X data to enhance the perception performance. Specifically, multi-agent sensing information is aggregated to form a holistic view and a novel proxy task is formulated to reconstruct the LiDAR point clouds across multiple connected agents to better reason multi-agent spatial correlations. Besides, we develop a V2X bird-eye-view (BEV) guided masking strategy which effectively allows the model to pay attention to 3D features across heterogeneous V2X agents (i.e., vehicles and infrastructure) in the BEV space. Noticeably, such a masking strategy effectively pretrains the 3D encoder with a multi-agent LiDAR point cloud reconstruction objective and is compatible with mainstream cooperative perception backbones. Our approach, validated through extensive experiments on representative datasets (i.e., V2X-Real, V2V4Real, and OPV2V) and multiple state-of-the-art cooperative perception methods (i.e., AttFuse, F-Cooper, and V2X-ViT), leads to a performance boost across all V2X settings.  Notably, CooPre achieves a 4\% mAP improvement on V2X-Real dataset and surpasses baseline performance using only 50\% of the training data, highlighting its data efficiency. Additionally, we demonstrate the framework's powerful performance in cross-domain transferability and robustness under challenging scenarios. The code will be made publicly available at \href{https://github.com/ucla-mobility/CooPre}{https://github.com/ucla-mobility/CooPre}.

\end{abstract}
%%%%%%%%%%%%%%%%%%%%%%%%%%%%%%%%%%%%%%%%%%%%%%%%%%%%%%%%%%%%%%%%%%%%%%%%%%%%%%%%

\section{Introduction}
Achieving autonomy in complex and open traffic environments poses significant challenges for single-vehicle vision systems. These systems often suffer from occlusions and a limited perception range due to each vehicle’s singular viewpoint, limiting the capacity of current deep learning approaches to develop a holistic 3D representation in the interacting environment. Vehicle-to-Everything (V2X) Cooperative Perception \cite{xiang2024v2x, xu_v2x-vit_2022, li_v2x-sim_2022, yu_v2x-seq_2023} emerges as a promising solution by providing each ego agent with a comprehensive understanding of the surrounding environment. Through collaboration among connected agents (vehicles or infrastructure), V2X facilitates the sharing of critical sensing information, thereby extending the perception range and mitigating occlusions. However, this paradigm introduces additional geometrical and topological information that the model must handle, necessitating the exploration of a robust representation that accounts for these elements. Obtaining such a representation often requires a large amount of annotated data, but the scale of current real-world V2X datasets \cite{xiang2024v2x, hao2024rcooper, yu_v2x-seq_2023, zimmer2024tumtrafv2x, openmars, xu_v2v4real_2023} is still limited compared to single-vehicle counterparts \cite{caesar_nuscenes_2020, chang_argoverse_2019,kitti, waymodataset}. Therefore, investigating the representation learning problem within the constraints of current V2X dataset scales is crucial. 

In multi-agent perception systems, each ego agent must learn a representation that manages complex agent interaction information to achieve effective cooperation. Specifically, in V2X scenarios, the ego agent must handle different sensor configurations among various agents, which operate at different ranges and placement positions. Capturing this underlying distribution is challenging solely through the supervision of hand-labeled 3D bounding boxes. This complexity poses a challenge for the "train-from-scratch" paradigm, as its learned representation heavily depends on the random initialization of model parameters and the quality and quantity of annotated 3D bounding boxes, thereby affecting overall performance. For example, as discussed in \cite{lu2023robust, wei2023asynchronyrobust, v2xadversial,xu_v2v4real_2023, cooperfuse}, such approach could be easily perturbed by synchronization or localization errors in real-world scenarios. DiscoNet \cite{disconet} offers an alternative paradigm by employing a teacher model to guide the student model during training. However, a significant amount of annotations is still required to train an effective teacher model that can facilitate the learning of the student model.

\begin{figure}[t]
  \centering
    \includegraphics[width=1.\linewidth]{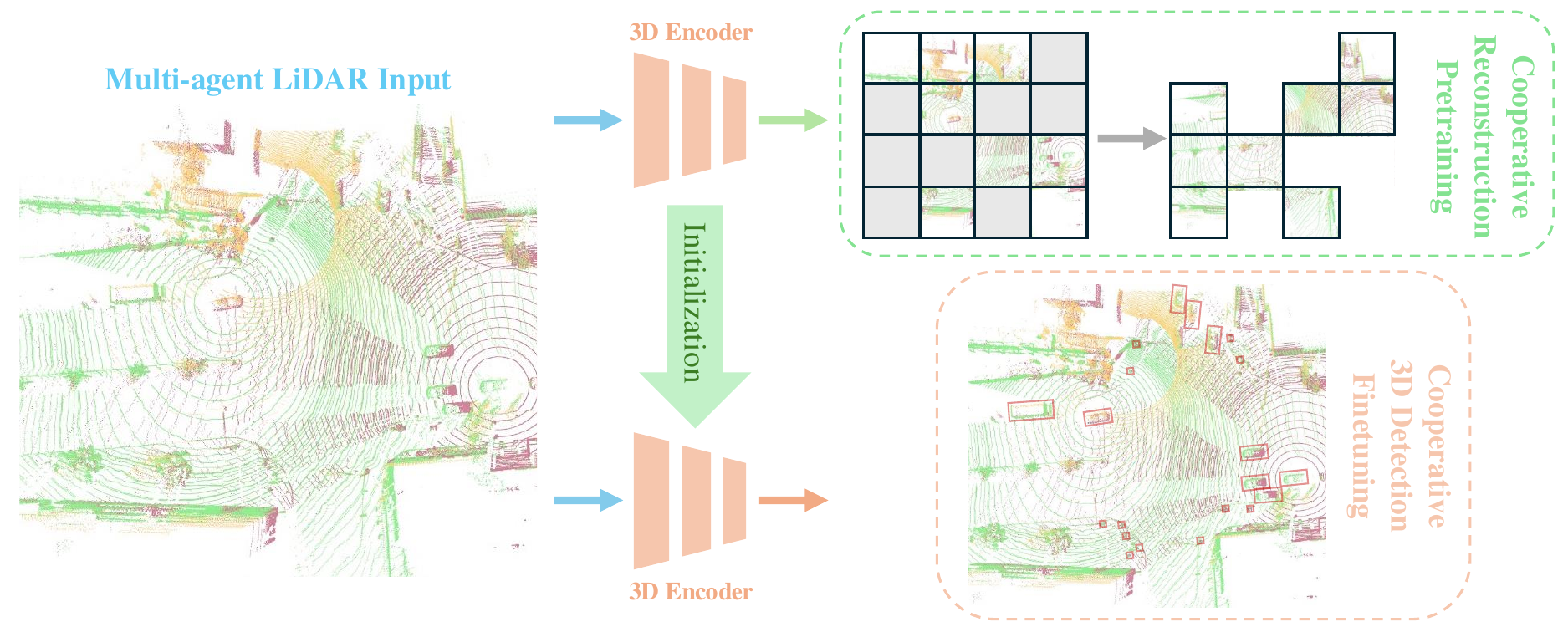}
    \caption{\textbf{Illustration of our CooPre framework.} Given multi-agent LiDAR input, CooPre pretrains the 3D encoder via LiDAR point cloud reconstruction. During the finetuning stage, the initialized 3D encoder will be used for 3D detection.}
    \vspace{-12pt}
  \label{fig:teaser}
\end{figure}

\begin{figure}[tbh]
  \centering
    \includegraphics[width= 1.\linewidth]{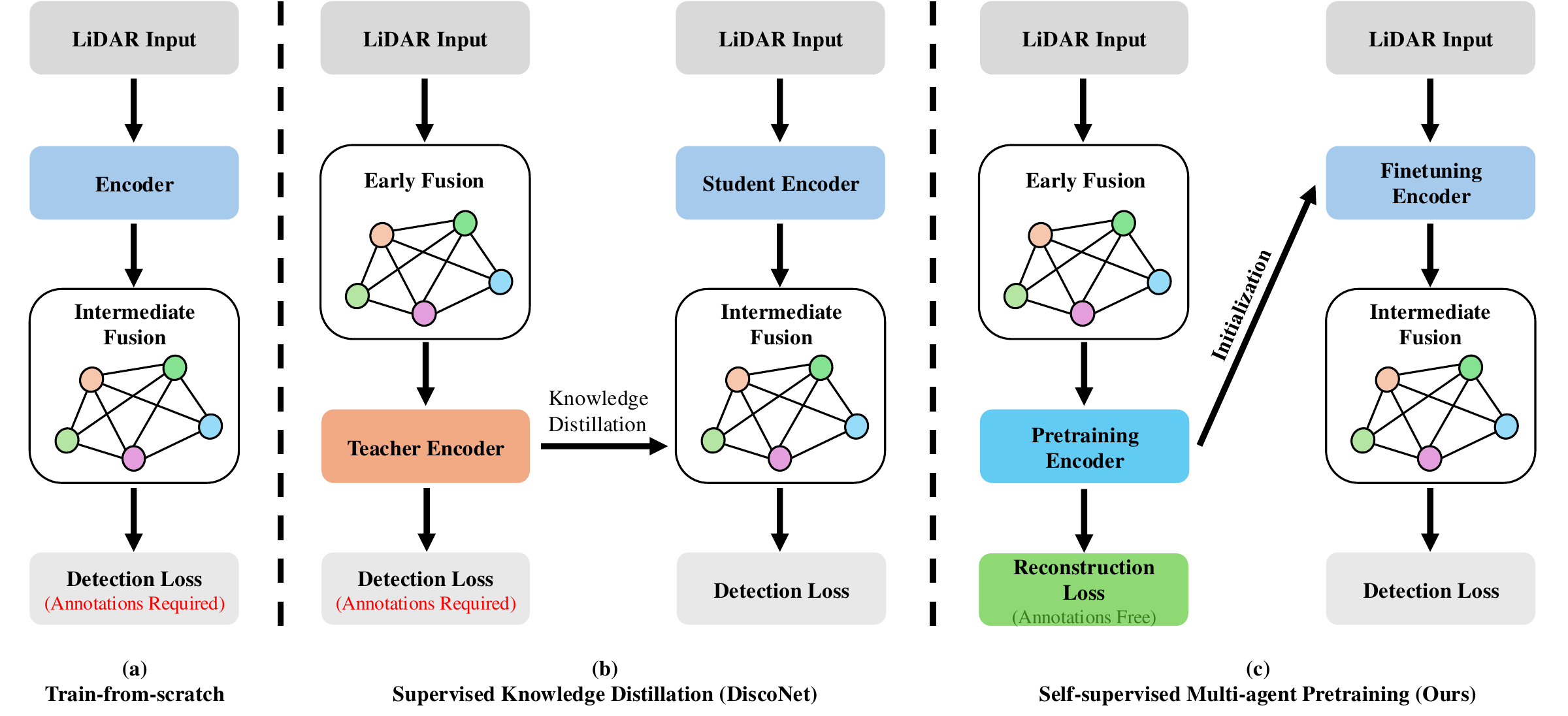}
    \caption{\textbf{Illustration of different training paradigms.} Compared with previous training paradigms, our paradigm uses multi-agent collaboration while being annotation-free.} 
  \label{fig:paradigm}
  \vspace{-12pt}
\end{figure}

To this end, we propose an effective self-supervised multi-agent \textbf{Coo}perative \textbf{Pre}training framework (\textbf{CooPre}) that enables the model to learn meaningful \textit{prior representation} of the holistic 3D environment before the perception task, as illustrated in Fig. \ref{fig:teaser}. Our framework leverages the benefits of unlabeled LiDAR point cloud data transmitted from different agents, allowing the model to reconstruct the point cloud location and learn essential prior knowledge of scenarios (e.g., intersections or corridors) and LiDAR sensor distributions (e.g., range, placement position, and sparsity) of each agent from a bird-eye-view (BEV) perspective. In the enlarged perception field, this heterogeneous reconstruction task also helps mitigate issues related to sparse feature points in far-range and occlusion scenarios. In addition, this method can be seamlessly integrated with state-of-the-art cooperative perception methods to improve perception capability. Compared to the other two training paradigms in Fig. \ref{fig:paradigm}, our pretraining framework is annotation-free, which is the main contribution to representation learning in V2X cooperative perception.

Through extensive experiments and ablation studies across three representative V2X datasets (i.e., V2X-Real \cite{xiang2024v2x}, V2V4Real \cite{xu_v2v4real_2023}, and OPV2V \cite{xu_opv2v_2022}) and three different V2X fusion methods \cite{xu_opv2v_2022, fcooper, xu_v2x-vit_2022}, we demonstrate the effectiveness of \textbf{CooPre} in V2X cooperative perception. Our findings show that: 1) \textbf{CooPre} boosts perception capabilities in scenarios involving occlusions and long-range perception, outperforming train-from-scratch and single-agent pretraining methods \cite{lin2024bevmae, bevcontrast}; 2) \textbf{CooPre} enables the model to learn important geometric and topological representations under heterogeneous sensor configurations, allowing better recognition of rigid-body objects like cars and trucks; 3) the learned representations from \textbf{CooPre} enhance data efficiency and demonstrate strong generalizability to unseen distributions, making them particularly advantageous for domain adaptation, especially in data scarcity scenarios. Moreover, our study underscores the critical role of well-learned 3D representations as a promising complement to task-specific design optimizations in V2X cooperative perception.

\section{Related Work}
\subsection{Cooperative Perception}
Single-vehicle systems struggle with occlusions and long-distance perception in complex traffic environments due to the limitation of the perception range of LiDAR devices. Cooperative systems, on the other hand, enhance detection performance by sharing raw data (Early Fusion), detection outputs (Late Fusion), or intermediate bird-eye-view (BEV) representations (Intermediate Fusion) among connected agents \cite{xu_opv2v_2022, wang_v2vnet_2020, hu2022where2comm}. With the recent advancement of a variety of simulation and real-world dataset curations \cite{xu_opv2v_2022, xu_v2v4real_2023, li_v2x-sim_2022, yu_dair-v2x_2022, yu_v2x-seq_2023, xiang2024v2x, hao2024rcooper}, many literature \cite{xu_opv2v_2022, xu_v2x-vit_2022, xu_cobevt_2022, wang_v2vnet_2020, fcooper, hu2022where2comm, yang2023how2comm, wu2024cmp} have been discussing the algorithmic designs of collaborative modes from vehicle-to-vehicle (V2V) collaboration to vehicle-to-everything (V2X) collaboration in different traffic scenarios. Noticeably, the intermediate fusion strategy has been the primary direction since it achieves the best trade-off between accuracy and bandwidth requirements. After applying a 3D encoder \cite{lang_pointpillars_2019, zhou_voxelnet_2018, second} to the input LiDAR feature, the intermediate fusion strategy involves projecting 3D features to BEV features where agents will perform interactions before final detection results. For example, AttFuse \cite{xu_opv2v_2022} utilizes a simple agent-wise single-head attention to fuse all features, whereas V2X-ViT \cite{xu_v2x-vit_2022} presents a unified vision transformer for robust multi-agent multi-scale perception. Despite their targeted designs, these methods all follow "train-from-scratch" paradigm and thus exhibit unstable performance when faced with V2X collaboration challenges, particularly due to sensor data heterogeneity issues \cite{xiang_hm-vit_2023, HEAL2024}. In this paper, we propose a model-agnostic pretraining framework to enhance the perception capability of these cooperative perception methods. 

\subsection{Lidar-based Self-supervised Learning}
Representation learning in autonomous driving \cite{forecastmae, trajmae, li2023pretraining, xu2022pretram, lin2024bevmae, min2023occupancymae, tian2023geomae, li2021metadrive, li2023scenarionet, bevcontrast} has been prevailingly investigated in single-vehicle systems. Stemming from the recent advancement of image reconstruction pretraining methods \cite{He2021MAE, Feichtenhofer2022STMAE}, point cloud pretraining reconstruction methods \cite{gao2022convmae, zhang2022point, yu2021pointbert, pang2022masked, huang2023ponder, zhu2023ponderv2, yang2023gdmae, lin2024bevmae, min2023occupancymae, tian2023geomae} are also proven effective in improving backbone model's robustness and generalizability. Recently, this approach has been applied to 3D representation learning of outdoor point clouds. Occupancy-MAE \cite{min2023occupancymae} applies a voxel-wise masking strategy to reconstruct masked voxels and predict occupancy. GD-MAE \cite{yang2023gdmae} proposes a multi-level transformer architecture and a multi-scale masking strategy with a lightweight generative decoder to recover masked patches. GeoMAE \cite{tian2023geomae} formulates the pretraining target to be geometric feature predictions, such as pyramid centroid, occupancy, surface normal, and curvature of point clouds. BEV-MAE \cite{lin2024bevmae} uses a BEV-guided masking strategy to learn BEV feature representations. Notably, while such pretraining methods have shown great potential in developing general feature representations, their improvement is limited due to the restricted perception field caused by perception range and occlusion in single-vehicle systems. Additionally, there is limited discussion on pretraining in multi-agent systems, with the most similar work being CORE \cite{coreICCV}, which proposes a BEV map reconstruction as an auxiliary task alongside the original perception target. Our work, on the other hand, provides the well-learned BEV representations that facilitate perception performance without additional computation efforts during the finetuning stage.

\begin{figure*}[tbh]
  \vspace{6pt}
  \centering
    \includegraphics[width= .95\linewidth]{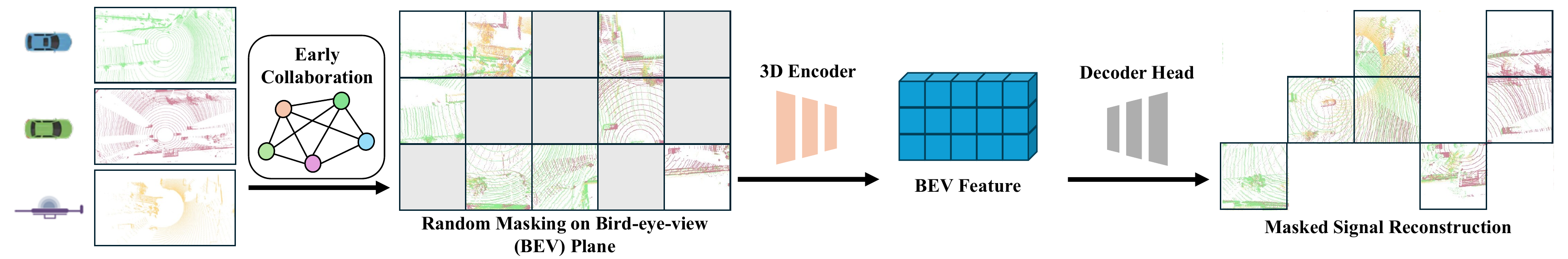}
    
    \caption{\textbf{Detailed pipeline of our proposed CooPre framework.} During the pretraining stage, the 3D encoder would be asked to perform a cooperative reconstruction task utilizing the vast amount of unlabeled LiDAR point cloud data transmitted from different agents. The masking strategy will be applied in a BEV-guided manner to take into account the reconstruction of far-range point clouds. } 
  \label{fig:pipeline}
  \vspace{-12pt}
\end{figure*}

\section{Method}
\subsection{Framework Overview}
Our CooPre framework, as illustrated in Fig. \ref{fig:pipeline}, is designed to facilitate 3D representation learning across heterogeneous V2X agents. To ensure robust cross-agent feature learning and mitigate feature sparsity issues, we propose three key design elements. First, we leverage early fusion to integrate LiDAR point cloud from collaborative agents, effectively expanding the ego agent’s perception field. This design mitigates the sparse feature issue and enables the subsequent LiDAR point cloud reconstruction objective to be applied to both the ego and collaborative agents. Second, instead of reconstructing the point cloud in voxel space, we reconstruct masked information on the BEV plane. This design enhances the model's feature learning process from a BEV perspective. Third, we require the model to reconstruct masked LiDAR point clouds from both the ego and collaborative agents, allowing the 3D encoder to learn the multi-agent LiDAR sensor distribution regardless of the ego agent’s perspective. Notably, our pretraining operates in BEV representation space, making it compatible with mainstream cooperative fusion methods \cite{xu_opv2v_2022, fcooper, xu_v2x-vit_2022} and widely used 3D encoder backbones \cite{second, lang_pointpillars_2019} during finetuning.

\subsection{Multi-agent Reconstruction Pretraining}

\noindent \textbf{Early Fusion of Multi-agent LiDAR point cloud.} We formulate the process of early fusion of raw LiDAR point cloud data across V2X agents as follows. Suppose we have ego agent $A_{ego}$ and a set of $N$ cooperative agents $A_i$ for $i \in \{1 \ldots N\}$ within the communication range, where each agent could be either Connected Autonomous Vehicle (CAV) or infrastructure. During the pretraining stage, each cooperative agent $A_i$ shares LiDAR point clouds and metadata information, such as poses, extrinsics, and agent type, to $A_{ego}$. We assume the transmission of LiDAR point clouds and metadata is well-synchronized. Consequently, after projecting the point clouds of each cooperative agent to the ego agent's coordinate, the perception point cloud field of the ego agent $A_{ego}$ includes its own LiDAR point clouds $P_{ego}$ as well as the LiDAR point clouds $P_i$ from each cooperative agent $A_i$. We refer to the collection of projected LiDAR point clouds from cooperative agents as $P_{coop} = \bigcup_i P_i$. In order for the encoder to learn invariant features from raw point cloud data from each agent, we perform data augmentations, including scaling, rotation, flip and downsampling \cite{lang_pointpillars_2019}. These augmentations enhance the model's generalization capabilities during pretraining, allowing it to effectively accomplish the pretraining objectives under varying conditions and viewpoints. Thus, the collection of original and augmented multi-agent LiDAR point cloud becomes our pretraining data corpus.

\noindent \textbf{V2X BEV-guided Masking Strategy.} We design a V2X BEV-guided masking strategy to enhance feature learning from a BEV perspective. For each ego agent $A_{ego}$, we suppose the shape of obtained BEV features from raw LiDAR input is $X \times Y \times C$. According to this shape, we define a BEV plane of size $X \times Y$ to divide up the plane and obtain BEV grids. We ensure that the size of masked BEV grids matches the resolution of BEV features obtained from 3D encoder. Each BEV grid $g_{i, j}$ would be used to determine whether the points inside the grid would be masked or not. After that, for each LiDAR point $p_k \in P_{ego} \bigcup P_{coop}$, we project it onto a corresponding BEV grid $g_{i,j}$ based on $p_k$'s x, y coordinates. With the help of projected point clouds from cooperative agents, the number of empty BEV grids largely decreases. We then randomly apply a high masking ratio towards non-empty BEV grids, where the point clouds inside the grid will be masked. Note that for a single-agent approach \cite{lin2024bevmae}, this strategy often leads to numerous BEV grids with sparse point clouds, making self-supervision both difficult and imprecise. In contrast, multi-agent collaboration proves advantageous, as it enriches the sparsely populated BEV grids with point cloud data from other collaborative agents. 

\noindent \textbf{Decoder design.} The decoder design is flexible and independent from downstream task since only the 3D encoder would be taken for finetuning. Unlike \cite{yang2023gdmae}, we utilize a lightweight decoder with one convolution layer for training efficiency. The decoder would take unmasked point cloud as input and output a fixed number of LiDAR point clouds for the multi-agent reconstruction objective. 

\noindent \textbf{Multi-agent LiDAR point cloud Reconstruction Objective.} As the ground-truth number of point clouds varies in each masked BEV grid, we utilize the Chamfer distance loss as the learning objective to measure the difference between predicted point clouds $\hat{P}$ and ground-truth point clouds $P$. Specifically, suppose in the masked BEV grid $g_{i,j}$, the reconstruction loss is defined as the following equation:
\begin{equation}
    \resizebox{.9\hsize}{!}{$L_{rec}(\hat{P},P)=\frac{1}{|\hat{P}|}\sum\limits_{\hat{p_i}\in\hat{P}}\min\limits_{p_j\in P}\left \| p_i-\hat{p_j} \right \|_2^2+\frac{1}{|P|}\sum\limits_{p_i\in P}\min\limits_{\hat{p_j}\in\hat{P}}\left \| p_i-\hat{p_j} \right \|_2^2$}
\end{equation}
We enforce the model to reconstruct the LiDAR point cloud from both ego and cooperative agent $P_{ego}$ and $P_{coop}$. We will show in the ablation study that such design is crucial for far-range feature learning across heterogeneous agents.

\begin{table*}[h]
  \vspace{6pt}
  \caption{Performance on V2X-Real dataset.}
  \vspace{-8pt}
  \label{tab:v2xreal}
  \centering
  \resizebox{0.8 \linewidth}{!}{\begin{tabular}{l|c|cc|cc|cc|cc}
    \toprule[1.2pt]
    \multirow{2}{*}{Dataset} & \multirow{2}{*}{Method} &  \multicolumn{2}{c}{Car} & \multicolumn{2}{c}{Pedestrian} & \multicolumn{2}{c}{Truck} & \multicolumn{2}{c}{Mean} \\
    && AP0.3 & AP0.5 & AP0.3 & AP0.5 & AP0.3 & AP0.5 & AP0.3 & AP0.5 \\
    \midrule
    V2X-Real VC
    & No Fusion & 51.0 & 48.0 & 35.8 & 20.4 & 38.9 & 36.3 & 41.9 & 34.9 \\ 
    & Early Fusion & 60.2 & 58.2 & 36.2 & 21.0 & 45.3 & 43.3 & 47.2 & 40.8 \\
    & Late Fusion & 59.0 & 56.5 & 36.7 & 17.6 & 44.0 & 39.7 & 46.5 & 37.9 \\
    & AttFuse \cite{xu_opv2v_2022} & 67.7 & 65.8 & 45.0 & 27.6 & 55.5 & 52.0 & 56.1 & 48.5 \\
    & F-Cooper \cite{fcooper} & 64.9 & 62.7 & 45.5 & 26.7 & 39.8 & 35.1 & 50.1 & 41.5\\ 
    & V2X-ViT \cite{xu_v2x-vit_2022} & 63.3 & 61.2 & 42.2 & 24.8 & 38.4 & 36.0 & 48.0 & 40.0\\ \cmidrule(l){2-10} 
    & \textbf{CooPre} \tiny\text{(AttFuse)} & \textbf{71.5} \tiny \textcolor{ForestGreen}{+3.8} & \textbf{70.2} \tiny\textcolor{ForestGreen}{+4.4} & \textbf{46.9} \tiny\textcolor{ForestGreen}{+1.9} & \textbf{28.0} \tiny\textcolor{ForestGreen}{+0.4} & \textbf{61.9} \tiny\textcolor{ForestGreen}{+6.4} & \textbf{58.3} \tiny\textcolor{ForestGreen}{+6.3} & \textbf{60.1} \tiny\textcolor{ForestGreen}{+4.0} & \textbf{52.2} \tiny\textcolor{ForestGreen}{+3.7} \\ 
    & \textbf{CooPre} \tiny\text{(F-Cooper)} & \textbf{67.0} \tiny\textcolor{ForestGreen}{+2.1} & \textbf{65.7} \tiny\textcolor{ForestGreen}{+3.0} & 43.6 \tiny\textcolor{Gray}{-1.9} & 25.4 \tiny\textcolor{Gray}{-1.3} & \textbf{49.3} \tiny\textcolor{ForestGreen}{+9.5} & \textbf{44.9} \tiny\textcolor{ForestGreen}{+9.8} & \textbf{53.3} \tiny\textcolor{ForestGreen}{+3.2} & \textbf{45.3} \tiny\textcolor{ForestGreen}{+3.8}\\
    & \textbf{CooPre} \tiny\text{(V2X-ViT)} & \textbf{66.1} \tiny\textcolor{ForestGreen}{+2.8} & \textbf{64.5} \tiny\textcolor{ForestGreen}{+3.3} & \textbf{45.3} \tiny\textcolor{ForestGreen}{+3.1} & \textbf{27.7} \tiny\textcolor{ForestGreen}{+2.9} & \textbf{43.5} \tiny\textcolor{ForestGreen}{+5.1} & \textbf{40.1} \tiny\textcolor{ForestGreen}{+4.1} & \textbf{51.7} \tiny\textcolor{ForestGreen}{+3.7} & \textbf{44.1} \tiny\textcolor{ForestGreen}{+4.1}\\
    \midrule
    V2X-Real V2V
    & No Fusion & 49.8 & 47.7 & 43.6 & 28.7 & 38.3 & 37.0 & 43.9 & 37.8 \\ 
    & Early Fusion & 62.5 & 60.0 & 43.3 & 28.1 & 47.3 & 44.7 & 51.0 & 44.3 \\
    & Late Fusion & 51.6 & 50.0 & 44.6 & 29.1 & 37.1 & 35.8 & 44.4 & 38.3 \\
    & AttFuse \cite{xu_opv2v_2022} & 66.9 & 65.3 & 46.0 & 30.4 & 53.8 & 50.2 & 55.5 & 48.6 \\
    & F-Cooper \cite{fcooper} & 64.6 & 63.0 & 41.9 & 26.9 & 48.9 & 44.8 & 51.8 & 44.9 \\
    & V2X-ViT \cite{xu_v2x-vit_2022} & 62.6 & 58.7 & 47.5 & 30.5 & 49.3 & 47.0 & 53.1 & 45.4 \\ \cmidrule(l){2-10} 
    & \textbf{CooPre} \tiny\text{(AttFuse)} & \textbf{71.4} \tiny\textcolor{ForestGreen}{+4.5} & \textbf{70.1} \tiny\textcolor{ForestGreen}{+4.8} & \textbf{49.7} \tiny\textcolor{ForestGreen}{+3.7} & \textbf{33.7} \tiny\textcolor{ForestGreen}{+3.3} & \textbf{57.5} \tiny\textcolor{ForestGreen}{+3.7} & \textbf{55.5} \tiny\textcolor{ForestGreen}{+5.3} & \textbf{59.5} \tiny\textcolor{ForestGreen}{+4.0} & \textbf{53.1} \tiny\textcolor{ForestGreen}{+4.5} \\
    & \textbf{CooPre} \tiny\text{(F-Cooper)} & \textbf{68.6} \tiny\textcolor{ForestGreen}{+4.0} & \textbf{67.0} \tiny\textcolor{ForestGreen}{+4.0} & \textbf{42.0} \tiny\textcolor{ForestGreen}{+0.1} & 26.4 \tiny\textcolor{Gray}{-0.5} & \textbf{51.0} \tiny\textcolor{ForestGreen}{+2.1} & \textbf{48.1} \tiny\textcolor{ForestGreen}{+3.3} & \textbf{53.9} \tiny\textcolor{ForestGreen}{+2.1} & \textbf{47.2} \tiny\textcolor{ForestGreen}{+2.3} \\
    & \textbf{CooPre} \tiny\text{(V2X-ViT)} & \textbf{66.9} \tiny\textcolor{ForestGreen}{+4.3} & \textbf{64.5} \tiny\textcolor{ForestGreen}{+5.8} & 46.7 \tiny\textcolor{Gray}{-0.8} & 29.2 \tiny\textcolor{Gray}{-1.3}& \textbf{54.1} \tiny\textcolor{ForestGreen}{+4.8} & \textbf{52.3} \tiny\textcolor{ForestGreen}{+5.3} & \textbf{55.9} \tiny\textcolor{ForestGreen}{+2.8} & \textbf{48.7} \tiny\textcolor{ForestGreen}{+3.3} \\
    \midrule
    V2X-Real IC
    & No Fusion & 55.7 & 48.6 & 35.2 & 20.4 & 46.3 & 45.5 & 45.8 & 38.2 \\ 
    & Early Fusion & 73.6 & 66.2 & 45.0 & 25.7 & 49.9 & 47.4 & 56.2 & 46.4 \\
    & Late Fusion & 74.5 & 72.5 & 47.8 & 27.4 & 66.1 & 57.8 & 62.8 & 52.6 \\
    & AttFuse \cite{xu_opv2v_2022} &  84.5 & 82.1 & 61.0 & 40.6 & 59.9 & 59.1 & 68.5 & 60.6 \\
    & F-Cooper \cite{fcooper} & 79.1 & 76.2 & 57.0 & 35.8 & 52.0 & 45.7 & 62.7 & 52.6 \\
    & V2X-ViT \cite{xu_v2x-vit_2022} & 81.0 & 76.9 & 44.9 & 29.2 & 37.3 & 36.8 & 54.4 & 47.6 \\ \cmidrule(l){2-10} 
    & \textbf{CooPre} \tiny\text{(AttFuse)} & \textbf{86.2} \tiny\textcolor{ForestGreen}{+1.7} & \textbf{84.1} \tiny\textcolor{ForestGreen}{+2.0} & 60.5 \tiny\textcolor{Gray}{-0.5} & 39.3 \tiny\textcolor{Gray}{-1.3} & 61.5 \tiny\textcolor{ForestGreen}{+1.6} & \textbf{61.0} \tiny\textcolor{ForestGreen}{+1.9} & \textbf{69.4} \tiny\textcolor{ForestGreen}{+0.9} & \textbf{61.4} \tiny\textcolor{ForestGreen}{+0.8} \\
    & \textbf{CooPre} \tiny\text{(F-Cooper)} & \textbf{84.1} \tiny\textcolor{ForestGreen}{+5.0} & \textbf{80.7} \tiny\textcolor{ForestGreen}{+4.5} & \textbf{58.4} \tiny\textcolor{ForestGreen}{+1.4} & 33.9 \tiny\textcolor{Gray}{-1.9} & 51.8 \tiny\textcolor{Gray}{-0.2} & \textbf{50.7} \tiny\textcolor{ForestGreen}{+5.0} & \textbf{64.8} \tiny\textcolor{ForestGreen}{+2.1} & \textbf{55.1} \tiny\textcolor{ForestGreen}{+2.5} \\
    & \textbf{CooPre} \tiny\text{(V2X-ViT)} & \textbf{81.4} \tiny\textcolor{ForestGreen}{+0.4} & \textbf{77.3} \tiny\textcolor{ForestGreen}{+0.4}
    & \textbf{45.8} \tiny\textcolor{ForestGreen}{+0.9} & 29.1 \tiny\textcolor{Gray}{-0.1} & \textbf{37.4} \tiny\textcolor{ForestGreen}{+0.1} & \textbf{37.2} \tiny\textcolor{ForestGreen}{+0.4} & \textbf{54.8} \tiny\textcolor{ForestGreen}{+0.4} & \textbf{47.9} \tiny\textcolor{ForestGreen}{+0.3} \\
    \midrule
    V2X-Real I2I
    & No Fusion & 67.3 & 61.5 & 49.2 & 32.3 & 53.2 & 49.0 & 56.6 & 47.6 \\ 
    & Early Fusion & 69.3 & 63.4 & 51.9 & 32.5 & 55.5 & 52.1 & 58.9 & 49.4 \\
    & Late Fusion & 79.6 & 78.0 & 67.3 & 46.7 & 64.4 & 55.8 & 70.4 & 60.2 \\
    & AttFuse \cite{xu_opv2v_2022} & 83.4 & 81.9 & 67.0 & 44.8 & 63.6 & 62.8 & 71.3 & 63.2 \\
    & F-Cooper \cite{fcooper} & 85.1 & 82.0	& 66.1 & 44.3 & 51.3 & 50.8 & 67.5 & 59.0\\ 
    & V2X-ViT \cite{xu_v2x-vit_2022} &	82.0 & 79.5 & 68.9 & 45.4 & 59.0 & 57.0 & 69.9 & 60.6 \\ \cmidrule(l){2-10} 
    & \textbf{CooPre} \tiny\text{(AttFuse)} & \textbf{84.6} \tiny \textcolor{ForestGreen}{+1.2} & \textbf{82.7} \tiny\textcolor{ForestGreen}{+0.8} & 66.5 \tiny\textcolor{Gray}{-0.5} & 45.2 \tiny\textcolor{ForestGreen}{+0.4} & \textbf{65.2} \tiny\textcolor{ForestGreen}{+1.6} & 62.5 \tiny\textcolor{Gray}{-0.3} & \textbf{72.1} \tiny\textcolor{ForestGreen}{+0.8} & \textbf{63.5} \tiny\textcolor{ForestGreen}{+0.3} \\
    & \textbf{CooPre} \tiny\text{(F-Cooper)} & \textbf{85.4} \tiny\textcolor{ForestGreen}{+0.3} & \textbf{82.1} \tiny\textcolor{ForestGreen}{+0.1} & \textbf{66.6} \tiny\textcolor{ForestGreen}{+0.5} & \textbf{45.1} \tiny\textcolor{ForestGreen}{+0.8} & \textbf{51.9} \tiny\textcolor{ForestGreen}{+0.6} & \textbf{51.1} \tiny\textcolor{ForestGreen}{+0.3} & \textbf{68.0} \tiny\textcolor{ForestGreen}{+0.5} & \textbf{59.4} \tiny\textcolor{ForestGreen}{+0.4}\\
    & \textbf{CooPre} \tiny\text{(V2X-ViT)} & \textbf{84.0} \tiny\textcolor{ForestGreen}{+2.0} & \textbf{82.4} \tiny\textcolor{ForestGreen}{+2.9} & 66.9 \tiny\textcolor{Gray}{-2.0} & 44.4 \tiny\textcolor{Gray}{-1.0} & \textbf{59.1} \tiny\textcolor{ForestGreen}{+0.1} & \textbf{58.7} \tiny\textcolor{ForestGreen}{+1.7} & \textbf{70.0} \tiny\textcolor{ForestGreen}{+0.1} & \textbf{61.9} \tiny\textcolor{ForestGreen}{+1.3} \\
    \bottomrule[1.2pt]
\end{tabular}}
\vspace{-6pt}
\end{table*}

\begin{figure}[h]
\vspace{4pt}
  \centering
    \includegraphics[width=1.\linewidth]{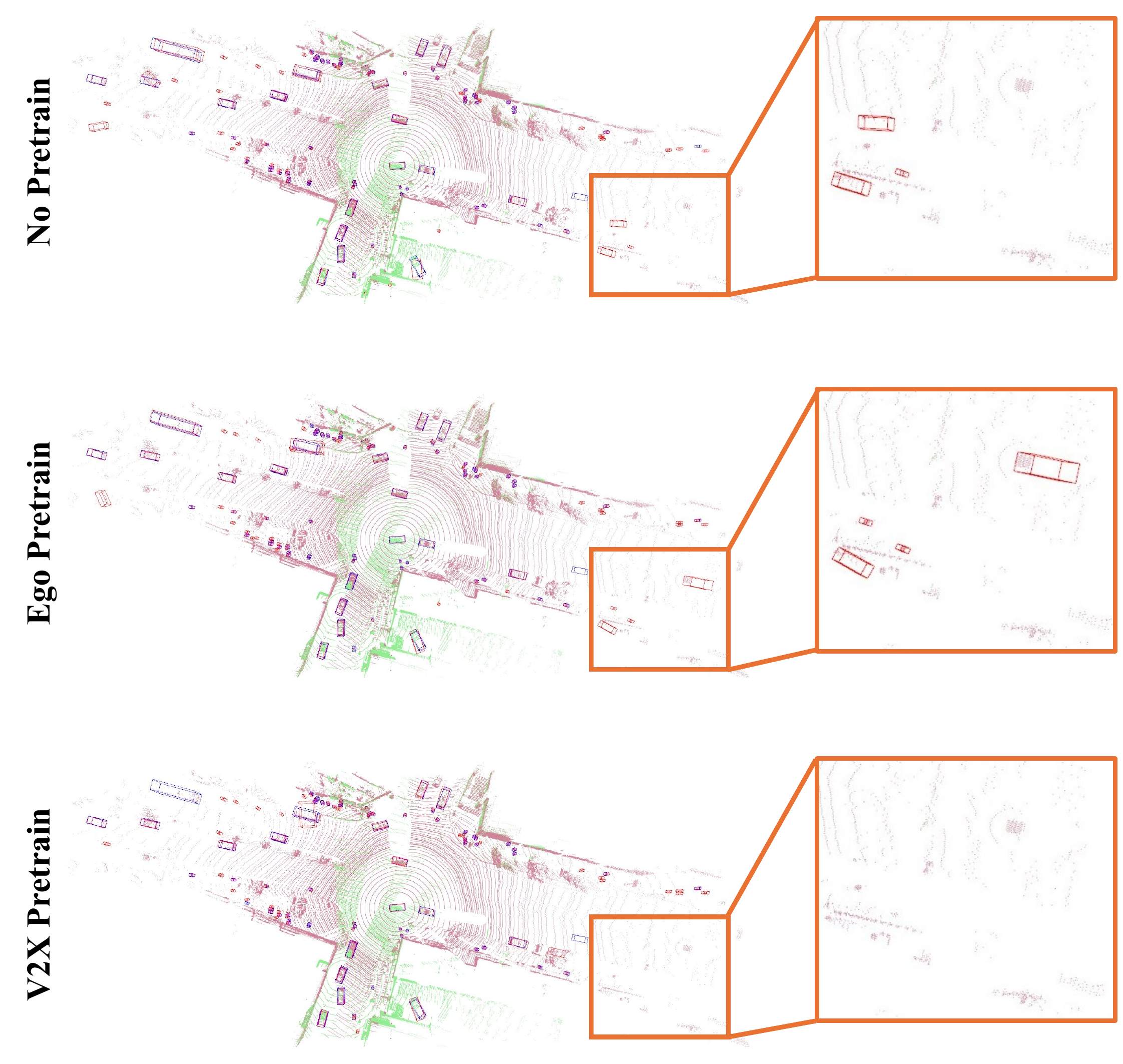}
    \caption{\textbf{Illustration of enhanced detection performance in long-range and occlusion scenarios.} Our multi-agent pretraining framework yields more robust perception capabilities in scenarios involving occlusions and long-range perception. Particularly, the occurrence of false perceptions significantly decreases in far-range scenarios where point cloud features are sparse. \textcolor{blue}{Blue} and \textcolor{red}{red} 3D bounding boxes correspond to the ground-truth and detection outputs, respectively. \textcolor{orange}{Orange} boxes denote the zoomed-in view of detection results.} 
  \label{fig:ablation_pretrainagents}
  % \vspace{-12pt}
\end{figure}

\begin{table*}[h]
  \caption{Performance on V2V4Real dataset.}
  \vspace{-8pt}
  \label{tab:v2v4real}
  \centering
  \resizebox{0.95 \linewidth}{!}{\begin{tabular}{l|cccc|cccc}
    \toprule[1.2pt]
    \multirow{2}{*}{Method} &  \multicolumn{4}{c}{Sync (AP0.5/0.7)} & \multicolumn{4}{c}{Async (AP0.5/0.7)}\\
    & Overall & 0-30m & 30m-50m & 50m-100m & Overall & 0-30m & 30m-50m & 50m-100m \\
    \midrule
    No Fusion & 52.6/36.2 & 76.6/58.7 & 45.9/28.7 & 10.0/5.0 & 52.6/36.2 & 76.6/58.7 & 45.9/28.7 & 10.0/5.0 \\
    Early Fusion & 66.7/34.5 & 83.3/48.9 & 51.0/25.3 & 48.4/19.9 & 59.5/28.7 & 82.5/46.2 & 43.5/21.1 & 29.8/9.0 \\
    Late Fusion & 69.9/40.2 & 81.4/44.0 & 60.7/38.5 & 56.9/36.2 & 63.9/32.5 & 78.9/40.3 & 56.8/32.9 & 39.4/16.8 \\
    AttFuse \cite{xu_opv2v_2022} & 71.2/44.2 & 87.3/55.5	& 55.6/39.8 & 51.1/25.7 & 63.8/34.9 & 85.7/51.4 & 47.8/31.0 & 33.4/10.5\\
    F-Cooper \cite{fcooper} & 70.6/41.8 & 84.4/52.8 & 59.9/37.8 & 51.5/26.2 & 62.2/34.1 & 82.5/50.0 & 51.7/30.9 & 31.9/10.7\\
    V2X-ViT \cite{xu_v2x-vit_2022} & 71.7/43.3 & 87.8/59.5 & 58.6/34.1 & 50.4/23.0 & 60.0/32.9 & 83.3/51.3 & 47.4/25.7 & 26.3/9.0 \\
    \midrule
    \textbf{CooPre} \tiny\text{(AttFuse)} & \textbf{74.3}/\textbf{49.3}  \tiny\textcolor{ForestGreen}{+3.1/+5.1} & \textbf{87.4}/\textbf{60.2} \tiny\textcolor{ForestGreen}{+0.1/+4.7} & \textbf{63.4}/\textbf{45.2} \tiny\textcolor{ForestGreen}{+7.8/+5.4} & \textbf{55.6}/\textbf{32.0} \tiny\textcolor{ForestGreen}{+4.5/+6.3} & \textbf{66.1}/\textbf{39.9} \tiny\textcolor{ForestGreen}{+2.3/+5.0} & \textbf{85.9}/\textbf{57.0} \tiny\textcolor{ForestGreen}{+0.2/+5.6} & \textbf{54.3}/\textbf{34.9} \tiny\textcolor{ForestGreen}{+6.5/+3.9} & \textbf{36.1}/\textbf{14.0} \tiny\textcolor{ForestGreen}{+2.7/+3.5}\\
    \textbf{CooPre} \tiny\text{(F-Cooper)} & \textbf{71.4}/\textbf{43.4} \tiny\textcolor{ForestGreen}{+0.8/+1.6} & \textbf{86.1}/\textbf{53.2} \tiny\textcolor{ForestGreen}{+1.7/+0.4} & \textbf{63.0}/\textbf{39.6} \tiny\textcolor{ForestGreen}{+3.1/+1.8} & 48.2/\textbf{29.5} \tiny\textcolor{Gray}{-3.3}\tiny\textcolor{ForestGreen}{/+3.3} & \textbf{63.7}/\textbf{34.8} \tiny\textcolor{ForestGreen}{+1.5/+0.7} & \textbf{84.7}/\textbf{50.3} \tiny\textcolor{ForestGreen}{+2.2/+0.3} & \textbf{55.0}/\textbf{31.1} \tiny\textcolor{ForestGreen}{+3.3/+0.2} & 29.7/\textbf{11.9} \tiny\textcolor{Gray}{-2.2}\tiny\textcolor{ForestGreen}{/+0.8}\\
    \textbf{CooPre} \tiny\text{(V2X-ViT)} & \textbf{73.1}/\textbf{43.9} \tiny\textcolor{ForestGreen}{+1.4/+0.6} & \textbf{88.9}/58.1 \tiny\textcolor{ForestGreen}{+1.1}\tiny\textcolor{Gray}{/-1.4} & \textbf{61.6}/\textbf{36.7} \tiny\textcolor{ForestGreen}{+3.0/+2.6} & \textbf{50.7}/\textbf{26.0} \tiny\textcolor{ForestGreen}{+0.3/+3.0} & \textbf{63.8}/\textbf{38.2} \tiny\textcolor{ForestGreen}{+3.8/+5.3} & \textbf{86.7}/\textbf{58.3} \tiny\textcolor{ForestGreen}{+3.4/+7.0} & \textbf{51.1}/\textbf{30.7} \tiny\textcolor{ForestGreen}{+3.7/+5.0} & \textbf{29.8}/\textbf{11.3} \tiny\textcolor{ForestGreen}{+3.5/+2.3} \\
  \bottomrule[1.2pt]
\end{tabular}}
\vspace{-12pt}
\end{table*}

\begin{table}[h]
  \caption{Performance on OPV2V dataset.}
  \vspace{-8pt}
  \label{tab:opv2v}
  \centering
  \resizebox{0.9 \linewidth}{!}{\begin{tabular}{l|cc|cc}
    \toprule[1.2pt]
    \multirow{2}{*}{Method} &  \multicolumn{2}{c}{Default} & \multicolumn{2}{c}{Culver}\\
    & AP0.5 & AP0.7 & AP0.5 & AP0.7 \\
    \midrule
    No Fusion & 71.3 & 60.4 & 64.6 & 51.7 \\
    Early Fusion & 87.7 & 81.3 & 82.1 & 73.8 \\
    Late Fusion & 84.6 & 77.5 & 80.8 & 68.2 \\
    AttFuse \cite{xu_opv2v_2022} & 89.3 & 82.6 & 87.5 & 76.0 \\
    F-Cooper \cite{fcooper} & 87.8 & 81.7 & 89.0 & 79.7 \\
    V2X-ViT \cite{xu_v2x-vit_2022} & 89.8 & 83.7 & 88.7 & 80.2\\
    \midrule
    \textbf{CooPre} \tiny\text{(AttFuse)} & \textbf{91.7} \tiny\textcolor{ForestGreen}{+2.4} & \textbf{86.6} \tiny\textcolor{ForestGreen}{+4.0} & \textbf{88.7} \tiny\textcolor{ForestGreen}{+1.2} & \textbf{80.0} \tiny\textcolor{ForestGreen}{+4.0}\\
    \textbf{CooPre} \tiny\text{(F-Cooper)} & \textbf{89.0} \tiny\textcolor{ForestGreen}{+1.2} & \textbf{83.4} \tiny\textcolor{ForestGreen}{+1.7} & \textbf{90.2} \tiny\textcolor{ForestGreen}{+1.2} & \textbf{82.0} \tiny\textcolor{ForestGreen}{+2.3} \\
    \textbf{CooPre} \tiny\text{(V2X-ViT)} & \textbf{91.9} \tiny\textcolor{ForestGreen}{+2.1} & \textbf{86.5} \tiny\textcolor{ForestGreen}{+2.8} & \textbf{89.9} \tiny\textcolor{ForestGreen}{+1.2} & \textbf{82.4} \tiny\textcolor{ForestGreen}{+2.2} \\
  \bottomrule[1.2pt]
\end{tabular}}
\vspace{-12pt}
\end{table}

\section{Experiments}
\subsection{Experiment Setting}
\noindent \textbf{Datasets.} We evaluate CooPre on three datasets, namely V2X-Real \cite{xiang2024v2x}, V2V4Real \cite{xu_v2v4real_2023}, and OPV2V \cite{xu_opv2v_2022}. V2X-Real is a large-scale, real-world V2X dataset that encompasses all V2X collaboration modes, including vehicle-centric (VC), infrastructure-centric (IC), vehicle-to-vehicle (V2V), and infra-to-infra (I2I). LiDAR sensors for this dataset were embodied with different sensor configurations and deployed in both intersection and corridor scenarios. Note that except for the V2V subset in the V2X-Real dataset, the rest of the subsets all exhibit such LiDAR sensor heterogeneity property. We also examine the effectiveness of CooPre on V2V4Real and OPV2V, two well-established benchmarks for real-world and simulated V2V cooperative perception.

\noindent \textbf{Evaluation Metrics.} Following previous evaluation protocols \cite{xiang2024v2x, xu_opv2v_2022, xu_v2v4real_2023}, we adopt the Average Precisions (AP) at the specified Intersection-over-Union (IoU) threshold as the metric to evaluate the detection performance. For V2X-Real \cite{xiang2024v2x} dataset, we evaluate the performance of different subclasses (i.e. car, pedestrian, and truck) and a final mean average precision (mAP) at the threshold of 0.3 and 0.5. For V2V4Real \cite{xu_v2v4real_2023} dataset, we evaluate the performance of the vehicle class at the threshold of 0.5 and 0.7 in both synchronized and asynchronized modes. For OPV2V \cite{xu_opv2v_2022} dataset, we evaluate the performance of the vehicle class at the threshold of 0.5 and 0.7 at their two separate test sets. \textcolor{ForestGreen}{Green} denotes the improvements over the corresponding backbone for all experiments.

\noindent \textbf{Implementation Details.} We evaluate CooPre using SECOND backbone \cite{second} since it obtains higher performance and faster training speed in the real-world dataset \cite{xiang2024v2x, xu_v2v4real_2023}. Note that CooPre could also be generalized to other 3D encoders such as PointPillar \cite{lang_pointpillars_2019}, as shown in Table \ref{tab:ablation_3dencoder}. All experiments are conducted in one Nvidia A6000 GPU. We employ AdamW \cite{kingma2014adam} optimizer with a weight decay of $1\times10^{-2}$ to optimize our models. During the pretraining stage, we train the model with a batch size of 4 for 15 epochs using a learning rate of 0.002, and we decay the learning rate with a cosine annealing \cite{cosineanneal}. We use a masking ratio of 0.7 in our main experiments and a fixed predicted point cloud number of 20. During the fine-tuning stage, the optimization process is identical to the train-from-scratch baselines. We finetune the model for 40 epochs in OPV2V dataset, 60 epochs in V2V4Real dataset, and 20 epochs in V2X-Real dataset. We also add normal point cloud data augmentations for all finetuning experiments, including scaling, rotation, and flip \cite{lang_pointpillars_2019}. Note that both pretraining and finetuning stages are conducted within the same dataset.

\begin{figure*}[tbh]
\vspace{8pt}
  \centering
    \includegraphics[width= 1.\linewidth]{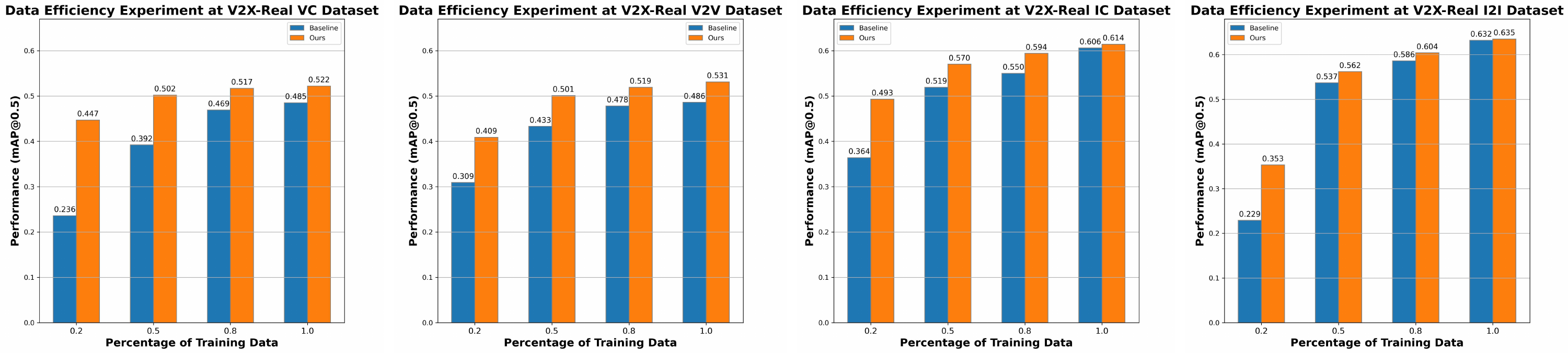}
    \caption{\textbf{Data efficiency experiment.} We evaluate the performance boost using CooPre under different ratios of finetuning data on V2X-Real dataset with AttFuse \cite{xu_opv2v_2022} as backbone. } 
  \label{fig:data_effiency}
  \vspace{-12pt}
\end{figure*}

\begin{table}[tb]
    \vspace{6pt}
    \centering
    \caption{Results of cross-domain transferability of our method (Metric: AP0.5 for Car Category).}
    \vspace{-8pt}
    \label{tab:transfer_learning}
    \resizebox{0.7 \linewidth}{!}{
    \begin{tabular}{l|cc}
        \toprule[1.2pt]
        \diagbox {Pretrain}{Finetune} & V2V4Real & V2X-Real V2V \\
        \midrule
        No Pretrain & 71.2 & 65.3 \\
        V2V4Real & 74.3 \tiny\textcolor{ForestGreen}{+3.1}  & 68.2 \tiny\textcolor{ForestGreen}{+2.9} \\
        V2X-Real V2V & 72.2 \tiny\textcolor{ForestGreen}{+1.0} & 70.1 \tiny\textcolor{ForestGreen}{+4.8} \\
        Combined & \textbf{74.5} \tiny\textcolor{ForestGreen}{+3.3} & \textbf{70.5} \tiny\textcolor{ForestGreen}{+5.2}\\
    \bottomrule[1.2pt]
    \end{tabular}}
\end{table}

\begin{figure}[t]
  \centering
    \includegraphics[width= 1.\linewidth]{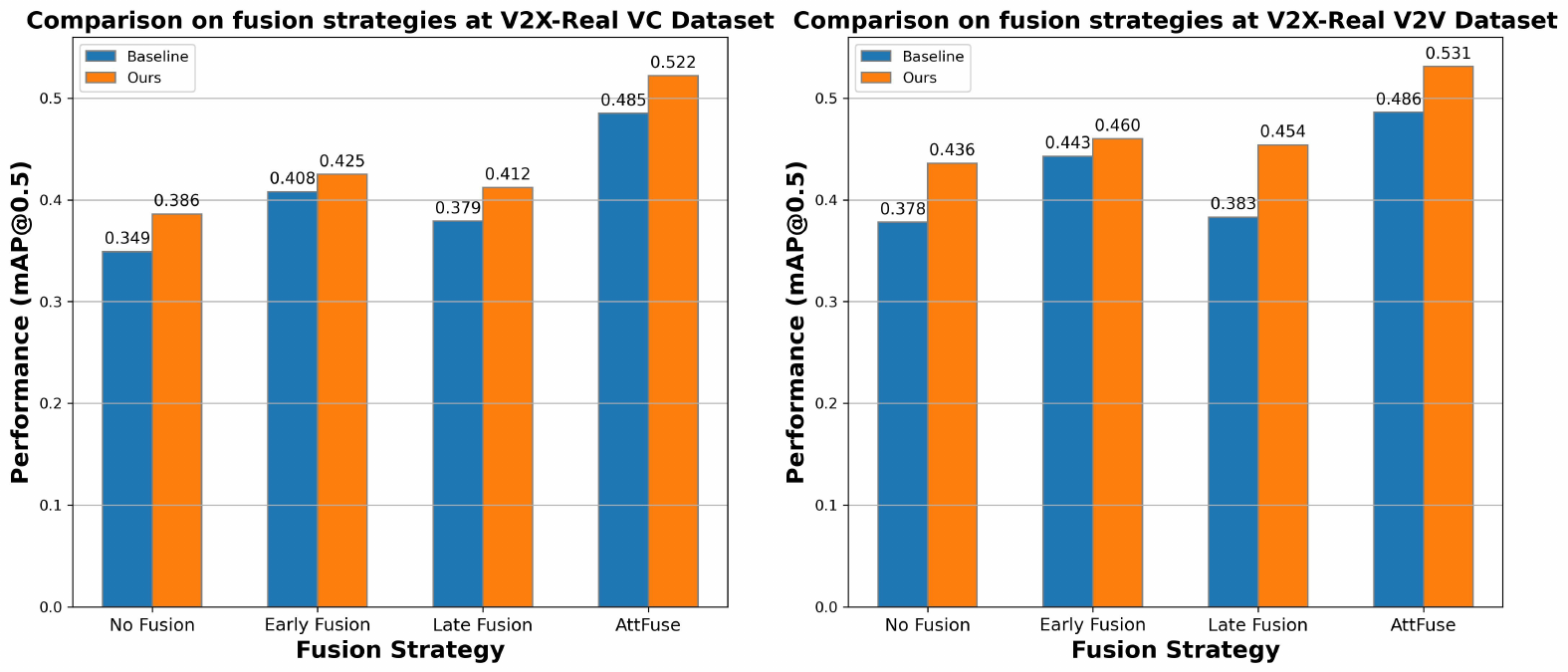}
    \caption{\textbf{Comparison of CooPre across different cooperative fusion strategies.} We assess the performance enhancement provided by CooPre under various fusion strategies on the V2X-Real VC and V2V datasets with AttFuse \cite{xu_opv2v_2022} as backbone. } 
  \label{fig:fusion}
  \vspace{-6pt}
\end{figure}

\begin{figure}[t]
  \centering
    \includegraphics[width= 1.\linewidth]{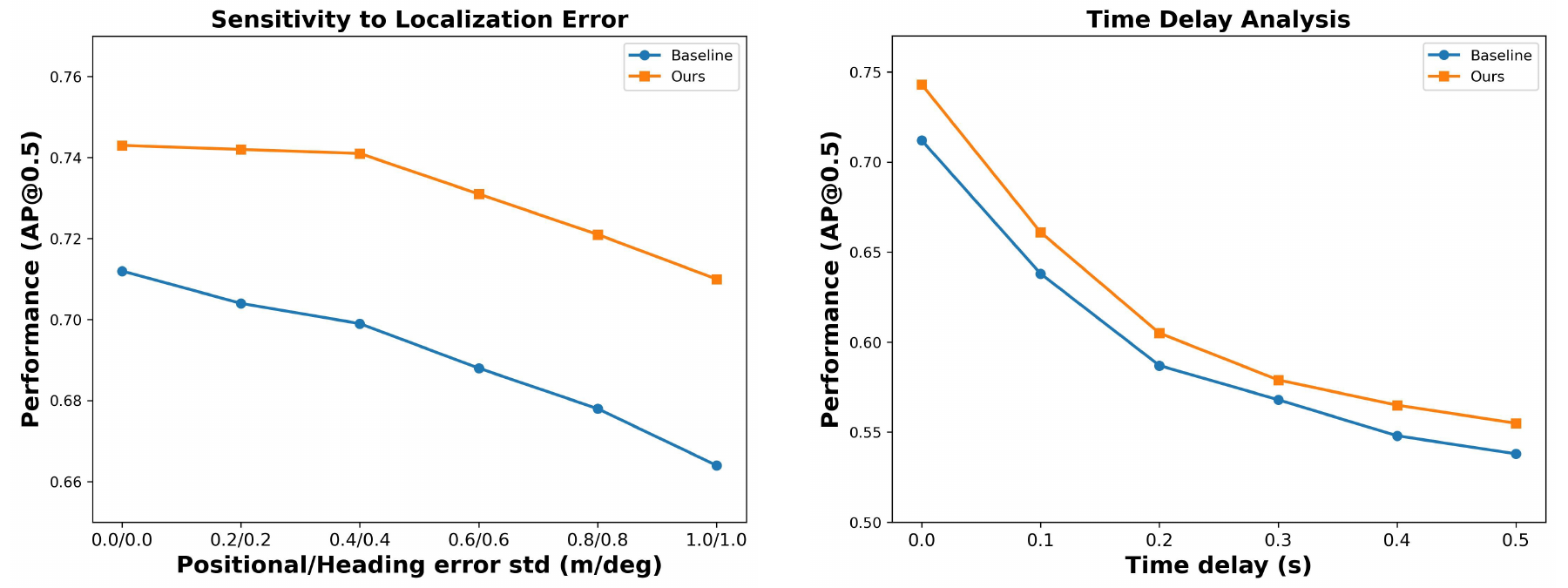}
    \caption{\textbf{Robustness assessment on localization errors and time delay.} Our method shows strong robustness against localization errors and time delay compared to the train-from-scratch baseline on V2V4Real dataset with AttFuse \cite{xu_opv2v_2022} as backbone.} 
  \label{fig:robustness}
  \vspace{-12pt}
\end{figure}

\begin{table}[t]
  \caption{Comparison with prior works (Metric: mAP0.5).}
  \vspace{-8pt}
  \label{tab:ablation_mae_agents}
  \centering
  \resizebox{.9 \linewidth}{!}{\begin{tabular}{l|c|c|c|c|c}
    \toprule[1.2pt]
    Dataset & Pretraining Method & Overall & 0-30m & 30-50m & 50-100m \\
    \midrule
    V2X-Real VC  
    & No Pretrain & 48.5 & 61.0 & 53.5 & 36.9 \\
    & BEVContrast \cite{bevcontrast} & 48.8 & 61.2 & 52.5 & 39.2  \\
    & Random Masking & 50.9 & 63.9 & 55.4 & 40.3 \\
    & CORE \cite{coreICCV} & 51.6 & 64.7 & 55.4 & 40.7 \\
    & STAR \cite{li2022multi} & 50.0 & 63.9 & 55.6 & 38.4 \\
    & BEV-MAE \cite{lin2024bevmae} & 51.1 & 64.5 & 55.7 & 39.4 \\
    & Ours & \textbf{52.2} & \textbf{64.9} & \textbf{55.9} & \textbf{42.1} \\
    \midrule
    V2X-Real V2V
    & No Pretrain & 48.6 & 64.2 & 52.3 & 36.0 \\
    & BEVContrast \cite{bevcontrast} & 46.3 & 63.1 & 54.1 & 32.9 \\
    & Random Masking & 50.9 & 65.5 & 54.7 & 40.4 \\
    & CORE \cite{coreICCV} & 51.9 & 65.9 & 55.0 & 41.8 \\
    & STAR \cite{li2022multi} & 51.2 & 66.0 & 54.4 & 40.6\\
    & BEV-MAE \cite{lin2024bevmae} & 50.5 & 65.4 & 54.6 & 38.7 \\
    & Ours & \textbf{53.1} & \textbf{66.1} & \textbf{55.4} & \textbf{42.7} \\
    
  \bottomrule[1.2pt]
\end{tabular}}
\vspace{-12pt}
\end{table}

\subsection{Main Results}
In Table \ref{tab:v2xreal}, we present the experimental results across all collaboration modes in V2X-Real \cite{xiang2024v2x} dataset. In the VC and V2V settings, CooPre substantially outperforms the train-from-scratch baselines across all classification categories. Using AttFuse \cite{xu_opv2v_2022} as an example, improvements are especially evident in the Car and Truck categories, with AP0.5 gains of 4.4 and 6.3 in the VC setting, and 4.8 and 5.3 in the V2V setting, respectively. This aligns with our pretraining design, which enables the model to learn more 3D geometrical and topological information beforehand, resulting in more accurate bounding box detection results under higher thresholds. On the other hand, our pretraining method shows an incremental effect on detecting pedestrians. We attribute this to two factors: 1) pedestrians are small-scale in nature and thus receive fewer LiDAR features than larger objects, and 2) pedestrians are non-rigid-body objects, making it more challenging for the model to learn their 3D features compared to rigid-body counterparts such as cars and trucks. On the infrastructure side, our pretraining method shows improvements for the Car and Truck categories, but we observe a subtle decrease in the Pedestrian category. We attribute this to the small-scale and non-rigid-body nature of pedestrians, which might be negatively affected by asynchronized results transmitted from connected vehicles, such as pose and localization errors \cite{lu2023robust, wei2023asynchronyrobust}. This asynchronization can alter the perception accuracy of static infrastructure observers. Such observation is also shown in the results of other cooperative methods \cite{fcooper, xu_v2x-vit_2022}.

We show the generalizability of CooPre in the V2V4Real \cite{xu_v2v4real_2023} and OPV2V \cite{xu_opv2v_2022} datasets, as shown in Table \ref{tab:v2v4real} and Table \ref{tab:opv2v}, respectively. For the V2V4Real dataset, after we pretrain and finetune the model, we test it under synchronized and asynchronized modes for a fair comparison following the original paper settings. The improvement is substantial in both testing settings. For the OPV2V dataset, we outperform the baseline by a large margin across all test sets. These results demonstrate the generalizability of CooPre across different domains with different fusion methods.

\subsection{Data Efficiency}
In this section, we investigate the benefits of our pretraining method in scenarios with limited labeled data using AttFuse \cite{xu_opv2v_2022} as backbone. Specifically, we randomly sample 20\%, 50\%, and 80\% of the training dataset and train the models with these annotated subsets. For our method, we pretrain the model on the entire training set and then finetune it on each sampled subset. As shown in Fig. \ref{fig:data_effiency}, our method outperforms train-from-scratch baselines across all settings. Notably, the performance gain of CooPre increases as the percentage of data decreases. Additionally, CooPre provides crucial guidance when collaboration involves different sensor configurations. For instance, in the V2X-Real VC dataset, the baseline method struggles to learn a good representation with limited annotations and the performance drops dramatically due to data heterogeneity issues. With our pretraining method, the model learns meaningful prior knowledge of different sensor distributions, leading to substantial improvements even when finetuning with less labeled data. These findings demonstrate the effectiveness of our method in data scarcity scenarios.

\subsection{Cross-domain Transferability}
We evaluate the cross-dataset transferability of our pretrained 3D encoder by finetuning it on another dataset using AttFuse \cite{xu_opv2v_2022} backbone. Since V2V4Real \cite{xu_v2v4real_2023} does not support multi-class classification, we only examine the perception results on car category which is shared by both domains. To isolate the effects of data heterogeneity, we investigate the performance across two real-world V2V datasets (V2X-Real V2V and V2V4Real), which have similar sensor configurations. As shown in Table \ref{tab:transfer_learning}, pretraining on a different domain improves performance compared to the train-from-scratch baseline. However, it performs worse than pretraining on the source domain due to the domain gap issue. We also conduct an experiment where we combine the V2X-Real V2V and V2V4Real datasets to create a large pretraining corpus. While this approach improves performance over pretraining on the source domain, the gains are less significant. This difference could be attributed to the scenario differences between the V2X-Real V2V and V2V4Real datasets. The former primarily focuses on intersection scenarios, whereas the latter includes a broader range of corridor scenarios in its training set.

\subsection{Ablation Studies and Discussion}
In this section, we conduct extensive experiments to explain how, where, and why our pretraining framework benefits current cooperative perception backbones.

\noindent \textbf{Comparison with other pretraining methods.} The comparison of pretraining methods is carried out in three aspects using AttFuse \cite{xu_opv2v_2022} as backbone. First, to evaluate reconstruction targets in a multi-agent pretraining setting, we adapt the reconstruction loss function from \cite{coreICCV, li2022multi} to align with our pretraining framework. Our approach demonstrates superior performance. Second, to analyze the impact of collaboration during pretraining, we compare our multi-agent pretraining method with the ego-agent pretraining approach from \cite{lin2024bevmae, bevcontrast}. As shown in Table \ref{tab:ablation_mae_agents}, multi-agent pretraining results in further performance gains. Additionally, both single-agent and multi-agent pretraining outperform the train-from-scratch approach. Lastly, we note variability in \cite{bevcontrast}, a contrastive learning based single-agent pretraining method, which may struggle to establish robust priors for multi-agent BEV representation compared to reconstruction-based pretraining methods.

\noindent \textbf{Extent of improvements in different perception ranges.} We also explore the enhancements across various perception ranges. As shown in Table \ref{tab:ablation_mae_agents} and Table \ref{tab:v2v4real}, the most substantial improvements are observed in middle and long ranges compared to the baselines. Since our multi-agent pretraining method enables a significantly larger perception field and provides supervision signals for feature learning on cooperative agents, it demonstrates greater robustness in handling long-range perception and occlusions compared to single-agent pretraining methods, as illustrated in Fig. \ref{fig:ablation_pretrainagents}.

\noindent \textbf{Improvements with different cooperative fusion strategies.} While our primary focus is on the intermediate fusion strategy, our method also applies to other fusion strategies, as depicted in Fig. \ref{fig:fusion}. We observe that our pretraining strategy benefits other cooperative fusion strategies as well, with the intermediate fusion strategy consistently demonstrating the best performance.

\noindent \textbf{Robustness assessment.} 
Following \cite{xu_v2x-vit_2022}, we evaluate our method's robustness towards localization error and time delay on the V2V4Real dataset, as illustrated in Fig. \ref{fig:robustness}. Our method is less sensitive to localization errors and shows strong robustness against time delay.

\noindent \textbf{Masking Ratio.} Table \ref{tab:ablation_mask_ratio} demonstrates the effect of the masking ratio within the range of 0.6 to 0.8. With a masking ratio of 0.7, CooPre achieves the best performance.

\noindent \textbf{Effectiveness on different 3D encoders.} Table \ref{tab:ablation_3dencoder} shows CooPre consistently improves the detection accuracy with different 3D encoders \cite{second, lang_pointpillars_2019}.

\begin{table}[tbh]
  \caption{Ablation on masking ratio.}
  \vspace{-8pt}
  \label{tab:ablation_mask_ratio}
  \centering
  \resizebox{1. \linewidth}{!}{\begin{tabular}{l|c|c|c|c}
    \toprule[1.2pt]
    Dataset & Method & Mask Ratio & mAP0.3 & mAP0.5 \\
    \midrule
    V2X-Real VC  &
    AttFuse \cite{xu_opv2v_2022}
    & 0.6/0.7/0.8 & 59.1/\textbf{60.1}/59.0 & 51.7/\textbf{52.2}/51.0 \\
    % \midrule
    % V2X-Real V2V
    % & 0.6/0.7/0.8 & 57.6/\textbf{59.5}/58.0 & 51.4/\textbf{53.1}/51.6  \\
  \bottomrule[1.2pt]
\end{tabular}}
\vspace{-12pt}
\end{table}

\begin{table}[tbh]
  \caption{Ablation on different 3D encoders in terms of mAP0.3/0.5.}
  \vspace{-8pt}
  \label{tab:ablation_3dencoder}
  \centering
  \resizebox{1.\linewidth}{!}{\begin{tabular}{l|c|cc|cc}
    \toprule[1.2pt]
    \multirow{2}{*}{Dataset} & 
    \multirow{2}{*}{Method} &
    \multicolumn{2}{c}{Second} & \multicolumn{2}{c}{PointPillar}\\
    && Original & W/ CooPre & Original & W/ CooPre \\
    \midrule
    V2X-Real VC &
    AttFuse \cite{xu_opv2v_2022} &
    56.1/48.5 & \textbf{60.1}/\textbf{52.2} & 42.5/33.8 & \textbf{44.0}/\textbf{35.2} \\
  \bottomrule[1.2pt]
\end{tabular}}
\vspace{-12pt}
\end{table}

\section{Conclusion}
We introduce \textbf{CooPre}, a multi-agent pretraining framework that prompts the representation to learn a holistic prior knowledge of the 3D environment before performing the perception task. The framework explores the intrinsic geometrical and topological information of scenarios and sensor distributions. Extensive experiments on representative datasets demonstrate the efficacy of the method as it outperforms the previous state-of-the-art methods in all V2X settings. Furthermore, we demonstrate this framework's strong generalizability, cross-domain adaptability, and data efficiency in cooperative perception. Future work includes extending this self-supervised cooperative pretraining paradigm to the field of cooperative perception and prediction task.

\section*{ACKNOWLEDGMENT}
This paper is partially supported by USDOT/FHWA Mobility Center of Excellence, National Science Foundation \# 2346267 POSE: Phase II: DriveX.

%%%%%%%%%%%%%%%%%%%%%%%%%%%%%%%%%%%%%%%%%%%%%%%%%%%%%%%%%%%%%%%%%%%%%%%%%%%%%%%%

%%%%%%%%%%%%%%%%%%%%%%%%%%%%%%%%%%%%%%%%%%%%%%%%%%%%%%%%%%%%%%%%%%%%%%%%%%%%%%%%
% \section*{APPENDIX}
% \textbf{Hyperparameter choice for synthetic data generation.} As mentioned in Section \ref{Synthetic}, the specific values of hyperparameters used in our mathematical formulation is indicated in Table \ref{tab:param_traj}.

% \section*{ACKNOWLEDGMENT}

% The preferred spelling of the word ÒacknowledgmentÓ in America is without an ÒeÓ after the ÒgÓ. Avoid the stilted expression, ÒOne of us (R. B. G.) thanks . . .Ó  Instead, try ÒR. B. G. thanksÓ. Put sponsor acknowledgments in the unnumbered footnote on the first page.

% %%%%%%%%%%%%%%%%%%%%%%%%%%%%%%%%%%%%%%%%%%%%%%%%%%%%%%%%%%%%%%%%%%%%%%%%%%%%%%%%

% References are important to the reader; therefore, each citation must be complete and correct. If at all possible, references should be commonly available publications.

{\small
\bibliographystyle{IEEEtran}
\bibliography{egbib}
}

\end{document}